\documentclass{article}
\usepackage{spconf,amsmath}
\usepackage{url}
\usepackage{booktabs}
\usepackage[dvipdfmx]{graphicx}
\newcommand{\bhline}[1]{\noalign{\hrule height #1}}  
 
\usepackage{multirow}
\usepackage{caption} 
\usepackage{subcaption}
\usepackage{color}

\title{LITE-HRNET PLUS: FAST AND ACCURATE FACIAL LANDMARK DETECTION}
\name{Sota Kato, Kazuhiro Hotta, Yuhki Hatakeyama, Yoshinori Konishi\address{Meijo University, SenseTime Japan}}

\begin{document}
%
\maketitle
\begin{abstract}
Facial landmark detection is an essential technology for driver status tracking and has been in demand for real-time estimations.
As a landmark coordinate prediction, heatmap-based methods are known to achieve a high accuracy, and Lite-HRNet can achieve a fast estimation.
However, with Lite-HRNet, the problem of a heavy computational cost of the fusion block, which connects feature maps with different resolutions, has yet to be solved.
In addition, the strong output module used in HRNetV2 is not applied to Lite-HRNet.
Given these problems, we propose a novel architecture called Lite-HRNet Plus.
Lite-HRNet Plus achieves two improvements: a novel fusion block based on a channel attention and a novel output module with less computational intensity using multi-resolution feature maps.
Through experiments conducted on two facial landmark datasets, we confirmed that Lite-HRNet Plus further improved the accuracy in comparison with conventional methods, and achieved a state-of-the-art accuracy with a computational complexity with the range of 10M FLOPs.
\end{abstract}
\section{Introduction}
Facial landmark detection is one of the key elements of face processing pipeline, such as the driver status tracking.
Automatic facial landmark detection has greatly benefited the development of such an agent-based interface \cite{zarkasi2022implementation,zhang2015adaptive}.
In recent years, a convolutional neural network (CNN) has achieved a high accuracy for facial landmark detection, and there are two types of approaches: regression \cite{feng2018wing} and heatmap-based approaches \cite{wu2018look}.
In particular, heatmap-based approaches using a high-resolution representation have achieved a high performance \cite{wang2020deep}.
However, in real-world environments, a fast estimation is also necessary for landmark detection in real-time using limited computational resources.
For our study, we set the target computational complexity of the model to be within the range of 10M FLOPs, which is the standard for processing in 10 milliseconds per image, because we must run facial landmark detection and other processes simultaneously in real-time on automotive SoCs, including face detection, head pose estimation, and gaze estimation. 
However, there are no methods that maintain high accuracy within this range of FLOPs in previous studies.

Motivated by the above problems, a novel network architecture called Lite-HRNet Plus was studied.
Figure 1 shows an overview of Lite-HRNet Plus.
Lite-HRNet Plus is based on Lite-HRNet \cite{yu2021lite}, which applies a shuffle block \cite{ma2018shufflenet} to HRNet \cite{sun2019deep} and introduces an efficient conditional channel weighting. 

We propose two novel architectures for Lite-HRNet Plus. 
First, we propose a stepped channel attention fusion (SCAF) block, which significantly reduce the computational complexity while preventing a decrease in accuracy.
Although Lite-HRNet has a fusion block that connects the feature maps between sub-networks of different resolutions, we confirmed that the heavily used point-wise convolutional layer in the fusion blocks is a computational bottleneck.
Our proposed SCAF block does not use point-wise convolutional layers and connects feature maps of different resolutions using the channel attention weights.  
By applying the SCAF block, we can reduce the entire computational complexity of the fusion block by approximately $80\%$.

\begin{figure}[t]
    \centering
    \includegraphics[scale=0.16]{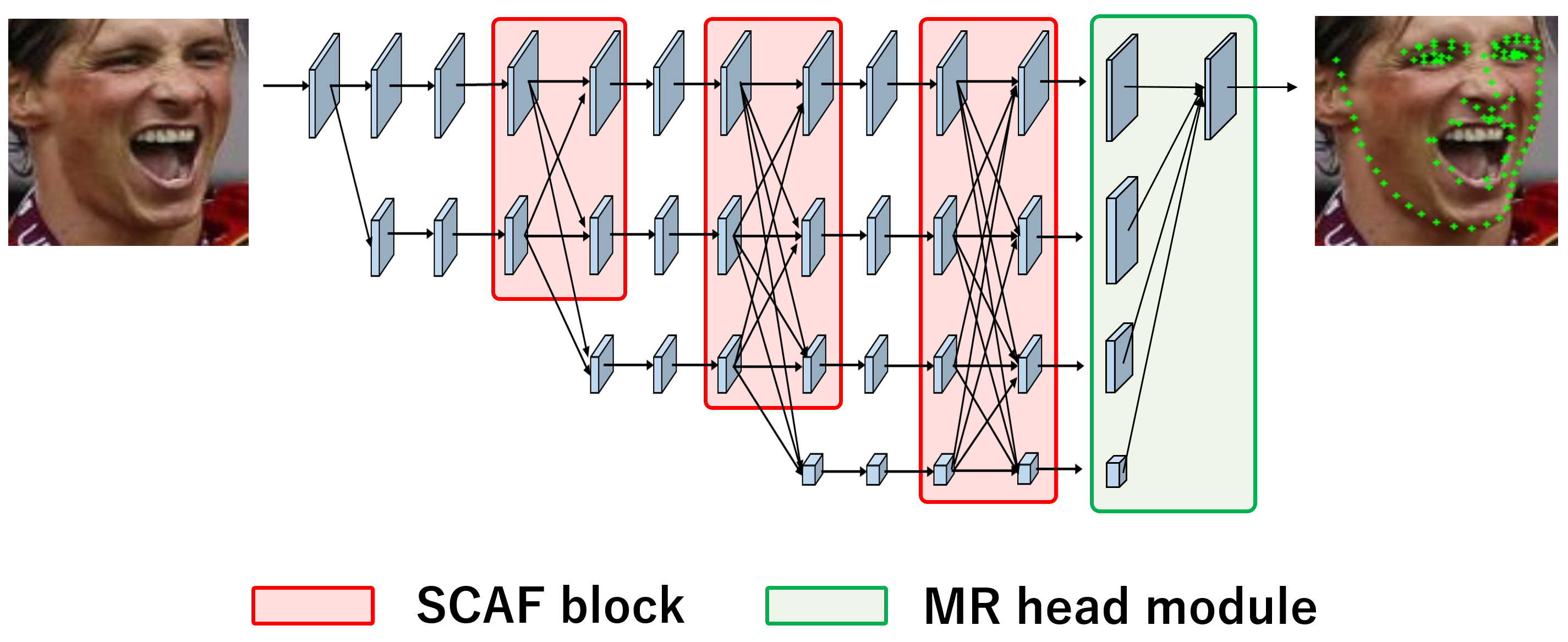}
    \caption{Overview of Lite-HRNet Plus. Our novel architectures, called the SCAF block and the MR head module, are applied to the original Lite-HRNet. }
\end{figure}

Second, we propose a multi-resolution (MR) output module, which uses feature maps of various resolutions for a prediction with a lower computational complexity.
Lite-HRNet adopts an HRNetV1 \cite{sun2019deep} based output module, which only outputs the high-resolution representation computed from the high-resolution stream. 
By contrast, HRNetV2 \cite{wang2020deep} adopts an output module that combines the representations from all high-to-low resolution parallel streams. 
Although the output module of HRNetV2 is effective for landmark detection \cite{wang2020deep}, it has a larger computational complexity than that of HRNetV1.
The MR output module combines feature maps of high-to-low resolution in a different way as HRNetV2, and can achieve a higher accuracy and a lower computational complexity than conventional output modules. 


We evaluated Lite-HRNet Plus experimentally on the facial landmark detection datasets \cite{wu2018look,sagonas2013300}. 
From the experiment results, we confirmed that Lite-HRNet Plus achieved significantly higher accuracy than conventional approaches, even if the computational complexity is within the range of 10M FLOPs.


\begin{figure}[t]
    \centering
    \includegraphics[scale=0.28]{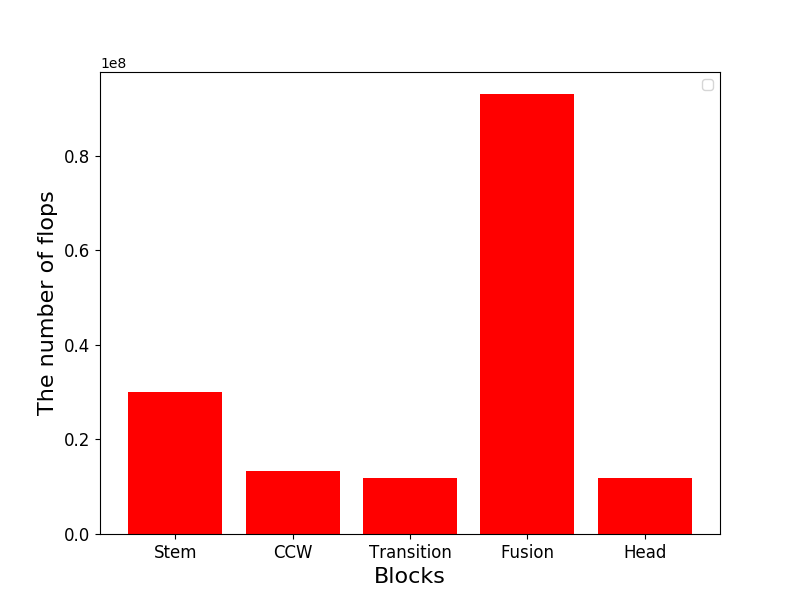}
    \caption{Analysis of the computational complexity of each block for Lite-HRNet. The vertical axis is the number of FLOPs and the horizontal axis is the block name.}
\end{figure}



\section{RELATED WORKS}
\begin{figure}[t]
    \centering
    \subfloat[]{\includegraphics[clip, width=3.3in]{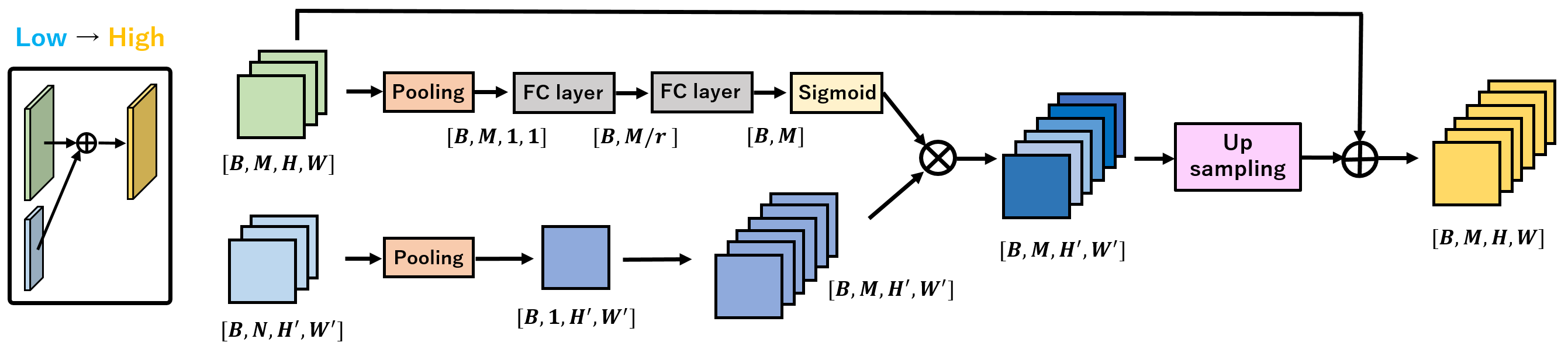}} \quad
    \vspace{0.2cm}
    \subfloat[]{\includegraphics[clip, width=3.3in]{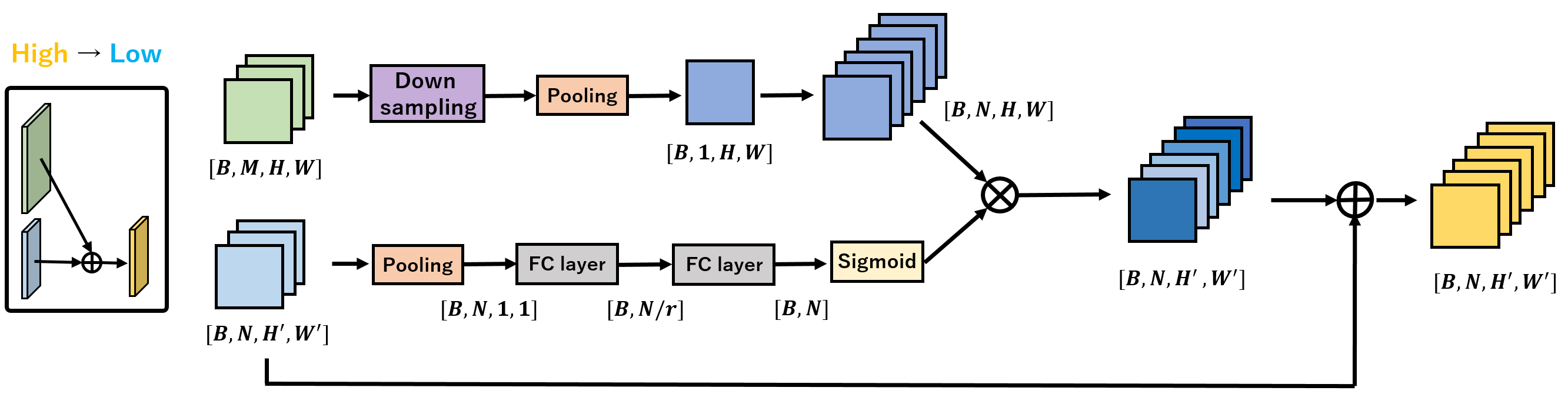}} \quad 
    \caption{Structures of SCAF block. The structure when (a) low-resolution feature maps are merged into high-resolution feature maps and (b) high-resolution feature maps are merged into low-resolution feature maps.
    }
    \label{fig:label-ABC}
\end{figure}


\textbf{Facial landmark detection.}
Facial landmark detection, which localizes pre-defined landmarks on human faces, is applicable to various tasks \cite{zhang2015pose,xia2019head}.
Traditionally, facial landmark detection using a CNN has been based on a regression model that directly estimates the coordinates of the landmarks \cite{feng2018wing,lv2017deep}.
However, in recent studies, models leveraging a CNN to achieve a heatmap regression are considered to be more accurate \cite{wu2018look}.
However, they have a problem in that the computational complexity is extremely large.

\noindent
\textbf{Lightweight convolutional neural networks.}
For image classification, MobileNets \cite{sandler2018mobilenetv2,howard2019searching} and ShuffleNets \cite{ma2018shufflenet,zhang2018shufflenet} are typical methods for lightweight networks.
MobileNet \cite{howard2017mobilenets} uses a depthwise and pointwise convolutional layer, and ShuffleNet \cite{zhang2018shufflenet} uses a group convolutional layer and a channel shuffle block to simplify the point-wise convolution.
For facial landmark detection, PIPNet \cite{jin2021pixel} is a lightweight method, and it can reduce the inference time largely without sacrificing the model accuracy.
Consequently, a state-of-the-art fast facial landmark detection is achieved.
For human pose estimation, Lite-HRNet \cite{yu2021lite} has been proposed as a strong, fast, and lightweight heatmap based method.
Lite-HRNet becomes a lightweight network by adding a shuffle block \cite{ma2018shufflenet} to HR-Net \cite{sun2019deep}, and by replacing the point-wise convolutional layer in the shuffle block with the attention layers. 

Although various lightweight methods have been proposed in facial landmark detection, there is a problem in that the accuracy deteriorates significantly as the computational complexity is reduced.
Although Lite-HRNet, which was originally proposed for human pose estimation, can be applied to facial landmark detection tasks and can alleviate the decrease in accuracy even when the computational complexity is reduced, it is insufficiently accurate when the computational complexity is constrained within 10M FLOPs range, which is the goal we are trying to achieve in this study.
Our proposed Lite-HRNet Plus adopts a new fusion block and a new output structure and can achieve both a high-speed estimation and a high accuracy.


\section{METHODOLOGY}

\subsection{Stepped channel attention fusion block} 
Figure 2 shows an analysis of the computational complexity of each block for Lite-HRNet.
Lite-HRNet consists of five blocks called the stem, conditional channel weighting (ccw), transition, fusion, and head blocks.
As shown in Figure 2, we confirmed that the fusion block, which connects the feature maps between sub-networks of different resolutions, has the highest computational complexity.
We consider the point-wise convolutional layer (PW Conv) included in the fusion block has a large computational complexity because it handles a huge number of channels.
PW Conv is a convolution operation that considers only the channel direction.
The computational complexity of PW Conv depends on the number of channels, and is written as $NMHW$, where $N$ and $M$ are the channel numbers of input and output feature maps, and $H$ and $W$ are the height and width of the feature maps.
Because feature maps of different resolutions have a different number of channels, it is necessary to align them to the same number of channels using PW conv, which is expected to result in a large computational complexity.

Therefore, we propose replacing the original fusion block with a stepped channel attention fusion (SCAF) block, which does not use PW Conv. 
As shown in Figure 3 (a), when merging low-resolution feature maps to high-resolution feature maps, the global average pooling along the channel direction is executed for low-resolution feature maps, and an averaged feature map is generated.
In high-resolution feature maps, the global average pooling along the spatial direction is executed and only channel information is left, and thus normalized attention weights within the range of $[0,1]$ are generated using fully connected layers and the sigmoid function.
The averaged low-resolution feature maps are then multiplied by the attention weights generated from high-resolution feature maps, and new feature maps are generated.
Finally, the spatial sizes of the fusion block and high-resolution feature maps are aligned through an up-sampling, and these feature maps are added to produce the output feature maps of the fusion block.
When high-resolution feature maps are merged into low-resolution feature maps, as shown in Figure 3 (b), new feature maps are generated in the same way; in addition, attention weights are generated from the low-resolution feature maps and multiplied by the averaged feature maps generated from the high-resolution feature maps.
The new feature maps are then added to the low-resolution feature maps.

In the SCAF block, only the weighting in the channel direction is computed, and thus the theoretical computational complexity is written as $\frac{2M^2}{r}$, where $r$ is a reduction ratio and uses the same symbol as above. 
Because the computational complexity of the SCAF block is determined only through $M$ and $r$ as opposed to $NMHW$, which is the computational complexity of PW Conv, the SCAF block can effectively reduce the computational complexity.

\subsection{Multi-resolution output module} 
\begin{figure}[t]
    \centering
    \includegraphics[scale=0.26]{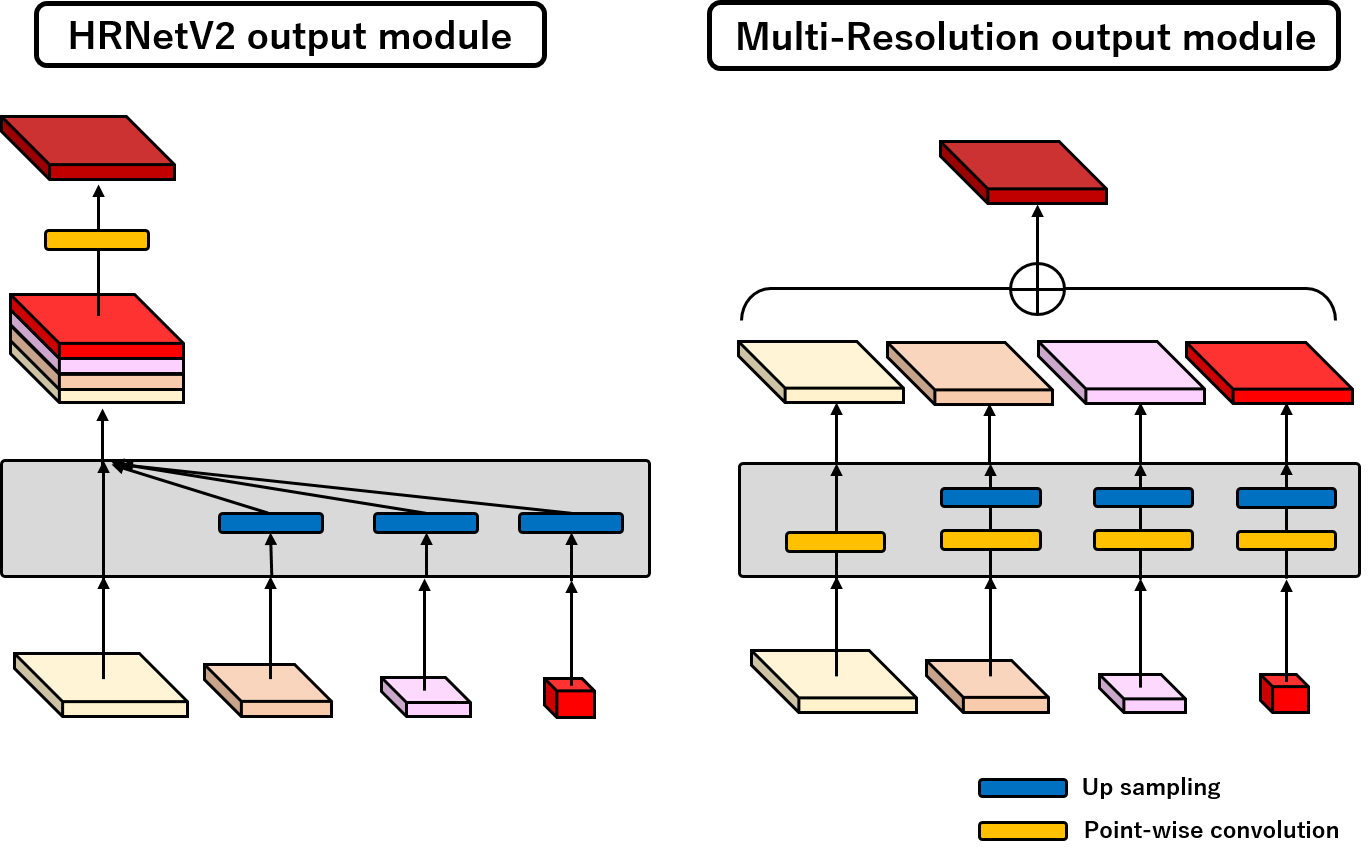}
    \caption{Comparison among the output modules.}
\end{figure}
\begin{table*}[t]
    \centering
    \caption{Evaluation results on facial landmark detection datasets using the NME metric.}
    \scalebox{0.75}{
    \begin{tabular}{lccccccccccccc}\bhline{0.9pt}
    &\multicolumn{7}{c}{\multirow{1}{*}{WFLW}}&\multicolumn{4}{c}{300W}&\multirow{3}{*}{\#Params}&\multirow{3}{*}{MFLOPs}\\
    \cmidrule(lr){9-12}%
    \cmidrule(lr){2-8}%
    \cmidrule(lr){9-11}%
    \multicolumn{0}{l}{Method}&All&Pose&Expr.&Illu.& Make.&Occu.&Blur&Com.&Chall.&All\,val.&All\,test.&&\\
    \bhline{0.5pt}
    $Large\:networks$ &&&&&&&&&&&\\
    \bhline{0.5pt}
    HRNetV2-W18  &4.99&8.49&5.25&4.89&5.27&5.92&5.37&4.42&6.61&4.85&5.37&9.66M&668.54\\
    \bhline{0.5pt}
    $Small\:networks$ &&&&&&&&&&&\\
    \bhline{0.5pt}
    MobileNetV2  &6.85&12.40&7.91&6.88&7.16&7.96&7.61&5.39&8.64&6.03&7.07&7.45M&256.80\\
    ShuffleNetV2  &6.43&11.23&7.31&6.26&6.98&7.57&6.99&5.20&8.24&5.80&6.74&6.66M&249.41\\
    Lite-HRNet&5.96&10.20&6.47&5.84&6.21&6.82&6.24&5.34&7.48&5.76&6.40&0.66M&33.61\\
    Dite-HRNet&6.81&11.02&7.53&6.59&6.98&7.57&7.07&5.22&7.48&5.66&6.27&0.70M&30.86\\
    Wing (MobileNetV2) &7.17&13.53&8.20&6.93&7.35&8.56&7.85&5.10&9.31&5.92&7.64&1.60M&41.56\\
    PIPNet (MobileNetV3)&5.87&10.79&\textbf{6.03}&5.50&5.93&7.19&6.50&4.07&7.10&4.66&5.58&4.47M&56.51\\
    PIPNet (MobileNetV3)&7.60&15.29&8.10&7.33&8.25&9.17&8.35&4.65&8.51&5.41&6.67&0.79M&11.46\\
    \cmidrule(r){1-14}%
    Lite-HRNet Plus (MSE loss)&5.76&10.14&6.41&5.63&6.03&6.76&6.17&4.78&7.07&5.23&5.89&1.51M&30.27\\
    Lite-HRNet Plus (MSE loss)&6.64&11.91&7.61&6.52&7.00&7.75&7.09&5.17&7.71&5.67&6.46&0.37M&10.25\\
    Lite-HRNet Plus (BCE loss)&\textbf{5.58}&\textbf{9.74}&6.13&\textbf{5.44}&\textbf{5.87}&\textbf{6.57}&\textbf{6.05}&\textbf{3.97}&\textbf{6.89}&\textbf{4.54}&\textbf{5.35}&1.51M&30.27\\
    Lite-HRNet Plus (BCE loss)&6.25&11.17&7.07&6.13&6.75&7.46&6.80&5.11&7.80&5.63&6.48&0.37M&10.25\\
    \bhline{0.9pt} 
    \end{tabular}
    }
\end{table*}
\begin{figure}[t]
    \centering
    \begin{tabular}{cc}
      \begin{minipage}{0.49\hsize}
        \centering
        \includegraphics[scale=0.28, angle=0]{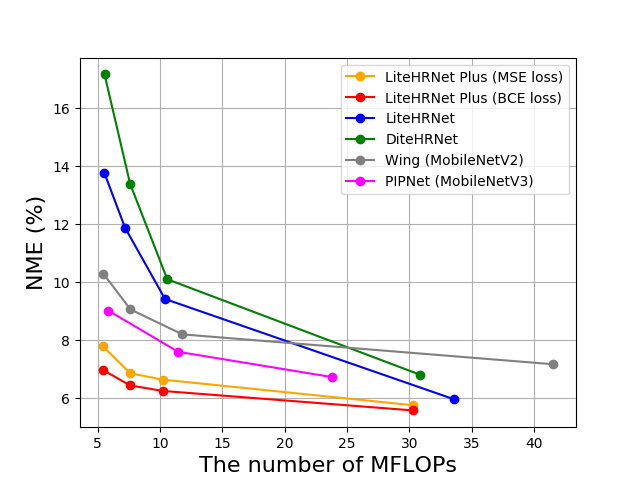}
        \subcaption{WFLW dataset}
    \end{minipage}%
    \begin{minipage}{0.52\hsize}
        \centering
        \includegraphics[scale=0.28, angle=0]{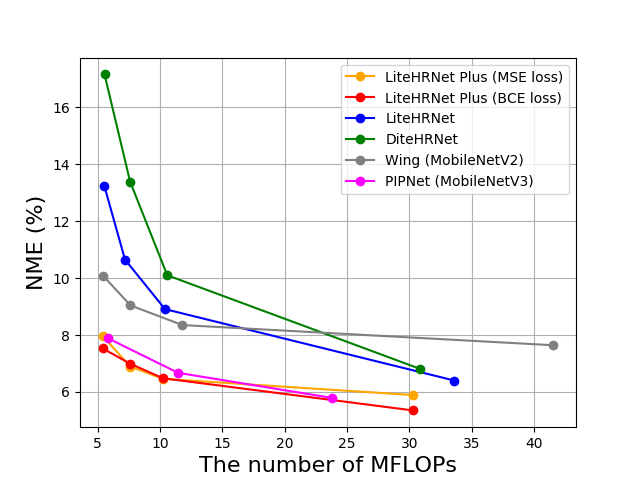}
        \subcaption{300W dataset}
    \end{minipage}%
    \end{tabular}
    \caption{Comparison among Lite-HRNet Plus and lightweight models in terms of the NME over FLOPs. 
    The closer a trade-off curve is to the bottom-left corner, the better its FLOPs-NME trade-off. 
    The existing methods were tested under the same environment as Lite-HRNet Plus.}
\end{figure}
In the HRNetV2 output module, low-resolution feature maps are up-sampled and concatenated with high-resolution feature maps, and the final prediction heatmaps whose channel number equals the number of landmarks are generated by PW Conv. 
Because this output module is highly effective in facial landmark detection \cite{wang2020deep}, we expect that it will also be effective for Lite-HRNet.
However, this module significantly increases the computational complexity for generating the final predictive heatmaps because it combines feature maps with different resolutions and outputs the heatmaps using PW Conv.

Therefore, we propose a multi-resolution (MR) output module.
Figure 4 shows a comparison of the structure of HRNetV2 and the MR output modules. 
In the MR output module, PW Conv is applied to each of the feature maps at different resolutions to output the prediction heatmaps, and each heatmap is added together to generate the final prediction.
We do not need to concatenate the feature maps, 
The computational complexity of the MR output module can be reduced to only approximately one-fifteenth times that of the HRNetV2 output module, and fitted in approximately 1.87 times the computational complexity of the Lite-HRNet output module.
\section{EXPERIMENTS}
\begin{table}[t]
    \centering
    \caption{Ablation studies of Lite-HRNet Plus.}
    \begin{tabular}{cc}
    \begin{minipage}{0.53\hsize}
        \centering
        \subcaption{Comparison among lightweight modules.}
        \scalebox{0.7}{\begin{tabular}{lcc}\bhline{0.7pt}
        Method&NME&MFLOPs\\
        \bhline{0.5pt}
        PW Conv &6.10&29.67\\
        Bottleneck&6.33&29.29\\
        Group Conv &5.99&29.31\\
        \bhline{0.5pt} 
        SCAF (ours) &\textbf{5.76}&30.27\\
        \bhline{0.7pt} 
        \end{tabular}
        }
    \end{minipage}%

    \begin{minipage}{0.45\hsize}
        \centering
    \subcaption{Inference speed.}
    \vspace{3.0mm}
    \scalebox{0.7}{
    \begin{tabular}{ccc}\bhline{0.7pt}
    MFLOPs&msec. / Image&FPS\\
    \bhline{0.5pt}
    30.27 &20.83&48.00\\
    10.25 &10.51&95.18\\
    \bhline{0.7pt} 
    \vspace{2.1mm}
    \end{tabular}
    }
    \end{minipage}%
    \end{tabular}
\end{table}

\subsection{Experimental setting} 
We used two datasets for facial landmark detection during the experiments.
The first dataset is the WFLW dataset \cite{wu2018look}, which consists of 7,500 face images for training and 2,500 face images for testing.
It is annotated with 98 landmarks and provides rich attribute labels such as large poses (Pose.), expressions (Expr.), illumination (Illu.), make-up (Make.), occlusions (Occu.), and blur (Blur). 
The second dataset is the 300W dataset \cite{sagonas2013300}, which is a widely adopted dataset for facial alignment with 3,148 images for training, 689 images for validation, and 600 images for testing.
It is manually annotated with 68 landmarks, and the validation data are divided into two sub-categories: 554 images for the common subset (Com.) and 135 images for the challenging subset (Chall.).

The batch size was set to 64, the number of training epochs was set to 210, and an Adam optimizer was used.
The initial learning rate was 0.002, and we used a learning rate schedule that decays the learning rate by 0.01 at the 170 epoch and again at the 200 epoch.
For the data pre-processing, training samples were resized to $96 \times 96$ pixels, randomly flipped horizontally, randomly scaled, randomly rotated with a selected angle of within $\theta$ = -$90^\circ$ to $90^\circ$, and normalized to 0-1 range.
For the inference, the images were resized to $96 \times 96$ pixels and normalized to 0-1 range.

As an evaluation metric, we applied the normalized mean error (NME), which uses the distance between the outer corners of the eyes as the reference distance to normalize the absolute errors. 

\subsection{Evaluation results} 
Table 1 shows the results of evaluation conducted on the WFLW and 300W datasets.
We compared our method with conventional lightweight approaches \cite{feng2018wing,yu2021lite,ma2018shufflenet,sandler2018mobilenetv2,jin2021pixel,LiZXZB22}.
Furthermore, although the mean squared error (MSE) loss is used as a loss function for conventional facial landmark detection \cite{wu2018look,jin2021pixel}, we also evaluated the binary cross entropy (BCE) loss because it is more sensitive to small errors than MSE loss.
As shown in Table 1, Lite-HRNet Plus performed better despite having the lowest computational complexity.
Comparing Lite-HRNet Plus with the BCE loss at 10.25M FLOPs to PIPNet at 11.46M FLOPs, the NME was improved by over $1.35\%$ in the WFLW dataset and the NME for test set was improved by $0.19\%$ in the 300W dataset.
These results demonstrate that Lite-HRNet Plus is superior to conventional lightweight methods.


Figure 5 shows a graphical comparison of the existing methods and our approach in terms of the speed and accuracy trade-off on the WFLW and 300W datasets. 
The closer the trade-off curve is the bottom-left corner, the better its FLOPs-NME trade-off. 
After verification using various lightweight methods, we confirmed that Lite-HRNet Plus achieved the best FLOPs-NME tradeoff.

Table 2 (a) shows a comparison among the lightweight techniques used for the fusion blocks.
We compared the SCAF block to other methods of reducing the computational complexity: PW Conv, PW Conv with a bottleneck (Bottleneck), and a group convolution (Group Conv). 
Bottleneck is a method for improving the computational efficiency proposed by \cite{sandler2018mobilenetv2}, and Group Conv is a method proposed in \cite{ma2018shufflenet}, which divides channels into some groups. 
As shown in Table 2 (a), the SCAF block can achieve the least computational complexity if the accuracy is the same, and can further contribute to a reduction of the computational complexity in comparison with conventional lightweight techniques.


Table 2 (b) shows the inference times of Lite-HRNet Plus
for a face image of 96$\times$96 pixels.
We measured the inference time per face image on an Intel Core i9-7960X CPU with the ONNX runtime \cite{onnxtoolkit} framework.
As shown in Table 2 (b), Lite-HRNet Plus achieved a rate of 48.00 fps for the 30.27M FLOPs model, and 95.18 fps for the 10.25M FLOPs model.
In conjunction with the results of Table 1, Lite-HRNet Plus can achieve a high accuracy at a real-time speed.

\section{CONCLUSION}

In this study, we proposed an effective framework called Lite-HRNet Plus for fast facial landmark detection.
As demonstrated through experiments conducted on two types of datasets, Lite-HRNet Plus showed a high accuracy in comparison with conventional lightweight methods despite the lower computational complexity, and it can run an inference at a real-time speed.
In addition, Lite-HRNet Plus may be effective in conducting other tasks, such as human pose estimation, which will be one of our future studies.

\bibliographystyle{IEEEbib}
\bibliography{strings,refs}

\begin{thebibliography}{10}

\bibitem{zarkasi2022implementation}
Ahmad Zarkasi, Fachrudin Abdau, Agung~Juli Anda, Siti Nurmaini, Deris Stiawan,
  Bhakti~Yudho Suprapto, Huda Ubaya, and Rizki Kurniati,
\newblock ``Implementation of facial landmarks detection method for face
  follower mobile robot,''
\newblock {\em Generic}, vol. 14, no. 1, pp. 19--24, 2022.

\bibitem{zhang2015adaptive}
Li~Zhang, Kamlesh Mistry, Ming Jiang, Siew~Chin Neoh, and Mohammed~Alamgir
  Hossain,
\newblock ``Adaptive facial point detection and emotion recognition for a
  humanoid robot,''
\newblock {\em Computer Vision and Image Understanding}, vol. 140, pp. 93--114,
  2015.

\bibitem{feng2018wing}
Zhen-Hua Feng, Josef Kittler, Muhammad Awais, Patrik Huber, and Xiao-Jun Wu,
\newblock ``Wing loss for robust facial landmark localisation with
  convolutional neural networks,''
\newblock in {\em Proceedings of the IEEE conference on computer vision and
  pattern recognition}, 2018, pp. 2235--2245.

\bibitem{wu2018look}
Wayne Wu, Chen Qian, Shuo Yang, Quan Wang, Yici Cai, and Qiang Zhou,
\newblock ``Look at boundary: A boundary-aware face alignment algorithm,''
\newblock in {\em Proceedings of the IEEE conference on computer vision and
  pattern recognition}, 2018, pp. 2129--2138.

\bibitem{wang2020deep}
Jingdong Wang, Ke~Sun, Tianheng Cheng, Borui Jiang, Chaorui Deng, Yang Zhao,
  Dong Liu, Yadong Mu, Mingkui Tan, Xinggang Wang, et~al.,
\newblock ``Deep high-resolution representation learning for visual
  recognition,''
\newblock {\em IEEE transactions on pattern analysis and machine intelligence},
  vol. 43, no. 10, pp. 3349--3364, 2020.

\bibitem{yu2021lite}
Changqian Yu, Bin Xiao, Changxin Gao, Lu~Yuan, Lei Zhang, Nong Sang, and
  Jingdong Wang,
\newblock ``Lite-hrnet: A lightweight high-resolution network,''
\newblock in {\em Proceedings of the IEEE/CVF Conference on Computer Vision and
  Pattern Recognition}, 2021, pp. 10440--10450.

\bibitem{ma2018shufflenet}
Ningning Ma, Xiangyu Zhang, Hai-Tao Zheng, and Jian Sun,
\newblock ``Shufflenet v2: Practical guidelines for efficient cnn architecture
  design,''
\newblock in {\em Proceedings of the European conference on computer vision
  (ECCV)}, 2018, pp. 116--131.

\bibitem{sun2019deep}
Ke~Sun, Bin Xiao, Dong Liu, and Jingdong Wang,
\newblock ``Deep high-resolution representation learning for human pose
  estimation,''
\newblock in {\em Proceedings of the IEEE/CVF conference on computer vision and
  pattern recognition}, 2019, pp. 5693--5703.

\bibitem{sagonas2013300}
Christos Sagonas, Georgios Tzimiropoulos, Stefanos Zafeiriou, and Maja Pantic,
\newblock ``300 faces in-the-wild challenge: The first facial landmark
  localization challenge,''
\newblock in {\em Proceedings of the IEEE international conference on computer
  vision workshops}, 2013, pp. 397--403.

\bibitem{zhang2015pose}
Zheng Zhang, Long Wang, Qi~Zhu, Shu-Kai Chen, and Yan Chen,
\newblock ``Pose-invariant face recognition using facial landmarks and weber
  local descriptor,''
\newblock {\em Knowledge-Based Systems}, vol. 84, pp. 78--88, 2015.

\bibitem{xia2019head}
Jiahao Xia, Libo Cao, Guanjun Zhang, and Jiacai Liao,
\newblock ``Head pose estimation in the wild assisted by facial landmarks based
  on convolutional neural networks,''
\newblock {\em Ieee Access}, vol. 7, pp. 48470--48483, 2019.

\bibitem{lv2017deep}
Jiangjing Lv, Xiaohu Shao, Junliang Xing, Cheng Cheng, and Xi~Zhou,
\newblock ``A deep regression architecture with two-stage re-initialization for
  high performance facial landmark detection,''
\newblock in {\em Proceedings of the IEEE conference on computer vision and
  pattern recognition}, 2017, pp. 3317--3326.

\bibitem{sandler2018mobilenetv2}
Mark Sandler, Andrew Howard, Menglong Zhu, Andrey Zhmoginov, and Liang-Chieh
  Chen,
\newblock ``Mobilenetv2: Inverted residuals and linear bottlenecks,''
\newblock in {\em Proceedings of the IEEE conference on computer vision and
  pattern recognition}, 2018, pp. 4510--4520.

\bibitem{howard2019searching}
Andrew Howard, Mark Sandler, Grace Chu, Liang-Chieh Chen, Bo~Chen, Mingxing
  Tan, Weijun Wang, Yukun Zhu, Ruoming Pang, Vijay Vasudevan, et~al.,
\newblock ``Searching for mobilenetv3,''
\newblock in {\em Proceedings of the IEEE/CVF international conference on
  computer vision}, 2019, pp. 1314--1324.

\bibitem{zhang2018shufflenet}
Xiangyu Zhang, Xinyu Zhou, Mengxiao Lin, and Jian Sun,
\newblock ``Shufflenet: An extremely efficient convolutional neural network for
  mobile devices,''
\newblock in {\em Proceedings of the IEEE conference on computer vision and
  pattern recognition}, 2018, pp. 6848--6856.

\bibitem{howard2017mobilenets}
Andrew~G Howard, Menglong Zhu, Bo~Chen, Dmitry Kalenichenko, Weijun Wang,
  Tobias Weyand, Marco Andreetto, and Hartwig Adam,
\newblock ``Mobilenets: Efficient convolutional neural networks for mobile
  vision applications,''
\newblock {\em arXiv preprint arXiv:1704.04861}, 2017.

\bibitem{jin2021pixel}
Haibo Jin, Shengcai Liao, and Ling Shao,
\newblock ``Pixel-in-pixel net: Towards efficient facial landmark detection in
  the wild,''
\newblock {\em International Journal of Computer Vision}, vol. 129, no. 12, pp.
  3174--3194, 2021.

\bibitem{LiZXZB22}
Qun Li, Ziyi Zhang, Fu~Xiao, Feng Zhang, and Bir Bhanu,
\newblock ``{Dite-HRNet}: Dynamic lightweight high-resolution network for human
  pose estimation,''
\newblock in {\em IJCAI-ECAI}, 2022.

\bibitem{onnxtoolkit}
``{ONNX: Open Neural Network Exchange},'' \url{https://github.com/onnx/onnx}.

\end{thebibliography}

\end{document}